\documentclass{article}

\usepackage[main,preprint]{neurips_2026}

\usepackage[utf8]{inputenc} 
\usepackage[T1]{fontenc}    
\usepackage{hyperref}       
\usepackage{url}            
\usepackage{booktabs}       
\usepackage{amsfonts}       
\usepackage{nicefrac}       
\usepackage{microtype}      
\usepackage{xcolor}         
\usepackage{amsmath}
\usepackage{bm}
\usepackage{amssymb}
\usepackage{mathtools}
\usepackage{amsthm}
\usepackage{array}
\usepackage{algorithm}
\usepackage{algpseudocode}
\usepackage{enumitem}
\usepackage{adjustbox}
\usepackage{siunitx}
\usepackage[table]{xcolor}

\theoremstyle{plain}
\newtheorem{theorem}{Theorem}[section]
\newtheorem{proposition}[theorem]{Proposition}
\newtheorem{lemma}[theorem]{Lemma}

\theoremstyle{definition}

\theoremstyle{remark}

\usepackage{multirow}
\algrenewcommand\algorithmicrequire{\textbf{Input:}}
\algrenewcommand\algorithmicensure{\textbf{Output:}}

\title{Attribution-Guided Continual Learning for Large Language Models}

\author{%
Yazheng Liu$^1$, Yuxuan Wan$^1$, Rui Xu$^1$, Xi Zhang$^2$, Sihong Xie$^1$, Hui Xiong$^1$\\
  $^1$ The Hong Kong University of Science and Technology (Guangzhou), Guangzhou, China\\
  $^2$ The Beijing University of Posts and Telecommunications, Beijing, China
}

\begin{document}

\maketitle

\begin{abstract}
Large language models (LLMs) often suffer from catastrophic forgetting in continual learning: after learning new tasks sequentially, they perform worse on earlier tasks. Existing methods mitigate catastrophic forgetting by data replay, parameter freezing, or regularization. However, these methods lack  understanding of LLM mechanisms and cannot distinguish which parameters store important knowledge from previous tasks and which  parameters can be updated for new tasks. To address this, we propose the attribution-guided continual fine-tuning framework that leverages Layer-wise Relevance Propagation (LRP) to estimate parameter importance based on the internal computational process of LLMs. During continual learning, parameters critical to previous tasks are constrained to receive smaller updates, while less relevant parameters remain available for learning new tasks. 
 Extensive experiments show that, compared with baseline methods, our approach reduces catastrophic forgetting while preserving adaptability to new tasks, highlighting the value of mechanistic attribution for continual fine-tuning of LLMs.
\end{abstract}
\section{Introduction}
Large Language Models (LLMs)~\cite{gpt4,llama,deepseek,comanici2025gemini,hui2024qwen2} have achieved exceptional performance across diverse tasks, such as multi-step reasoning\cite{wei2022chain,wang2022self}, instruction following~\cite{ouyang2022training,chung2024scaling}, and code geneartion~\cite{roziere2023code,chen2021evaluating}. 
However, LLMs are usually trained on static, broad-domain data. This can lead to performance degradation when the target domain changes~\cite{chen2023lifelong,dhingra2022time,lu2025fine}.
To adapt LLMs to downstream tasks while preserving prior knowledge, researchers  use continual learning method~\cite{van2022three,wang2024comprehensive}. 
Continual learning trains models on a sequence of tasks but faces a challenge known as catastrophic forgetting~\cite{mccloskey1989catastrophic,mcclelland1995there}: fine-tuning LLM on a new task can substantially degrade its performance on previously learned tasks.

Existing approaches have been proposed to  mitigate catastrophic forgetting in continual learning. Replay methods~\cite{sun2019lamol,huang2024mitigating,scialom2022fine,abbes2025revisiting} retain examples from previous tasks and train them together with current task data.
Regularization methods~\cite{zhang2023copr,rebuffi2024incremental,li2017learning} penalize large deviations from the previous model in parameter space.
Freezing methods fix selected LLM parameters to reduce forgetting~\cite{zheng2025spurious}. However, these methods lack a mechanistic understanding of how knowledge are distributed across LLM parameters. They may constrain parameters that are necessary for adapting to new tasks or modify parameters that are critical for retaining prior knowledge. 

To understand the internal mechanisms of LLMs in continual learning, we first explore the importance of parameters in LLMs on different tasks. Let $\theta$ denote the model parameters, $\mathcal{W}^{(l)}(\theta)$ be the set of model parameters in its $l$-th layer, and $W^{(l)} \in \mathcal{W}^{(l)}(\theta)$ represent an individual model parameter. For a task $\mathcal{T}$, we compute the importance of each parameter with respect to the next-token logits using the attribution method introduced in Section~\ref{sec:lrp_attribution_llms}. We then select the top-$K$ most important elements in $W^{(l)}$, denoted by
$\mathcal{P}_{K}\big(W^{(l)}; \theta, \mathcal{T}\big)$.
Given two tasks, $\mathcal{T}_1$ and $\mathcal{T}_2$, we compare independent single-task fine-tuning with sequential continual fine-tuning, as illustrated in Figure~\ref{fig:method}(a,b). In the single-task setting, a pretrained LLM is fine-tuned separately on each task, producing task-specific models $\theta_1^\prime$ and $\theta_2^\prime$. 
We compute $\mathcal{P}_{K}\big(W^{(l)}; \theta_1^\prime, \mathcal{T}_1\big)$ and $\mathcal{P}_{K}\big(W^{(l)}; \theta_2^\prime, \mathcal{T}_2\big)$. In the continual setting, the pretrained model is fine-tuned sequentially on $\mathcal{T}_1$ and $\mathcal{T}_2$, resulting in final parameters $\theta_2$.
 We obtain $\mathcal{P}_{K}\big(W^{(l)}; \theta_2, \mathcal{T}_1\big)$ and $\mathcal{P}_{K}\big(W^{(l)}; \theta_2, \mathcal{T}_2\big)$. Finally, we quantify the similarity of $\mathcal{P}_{K}\big(W^{(l)}; \theta_1^\prime, \mathcal{T}_1\big)$ and $\mathcal{P}_{K}\big(W^{(l)}; \theta_2^\prime, \mathcal{T}_2\big)$, $\mathcal{P}_{K}\big(W^{(l)}; \theta_2, \mathcal{T}_1\big)$ and $\mathcal{P}_{K}\big(W^{(l)}; \theta_2, \mathcal{T}_2\big)$.
 \begin{figure}[htbp]
  \centering
  \includegraphics[width=0.95\linewidth]{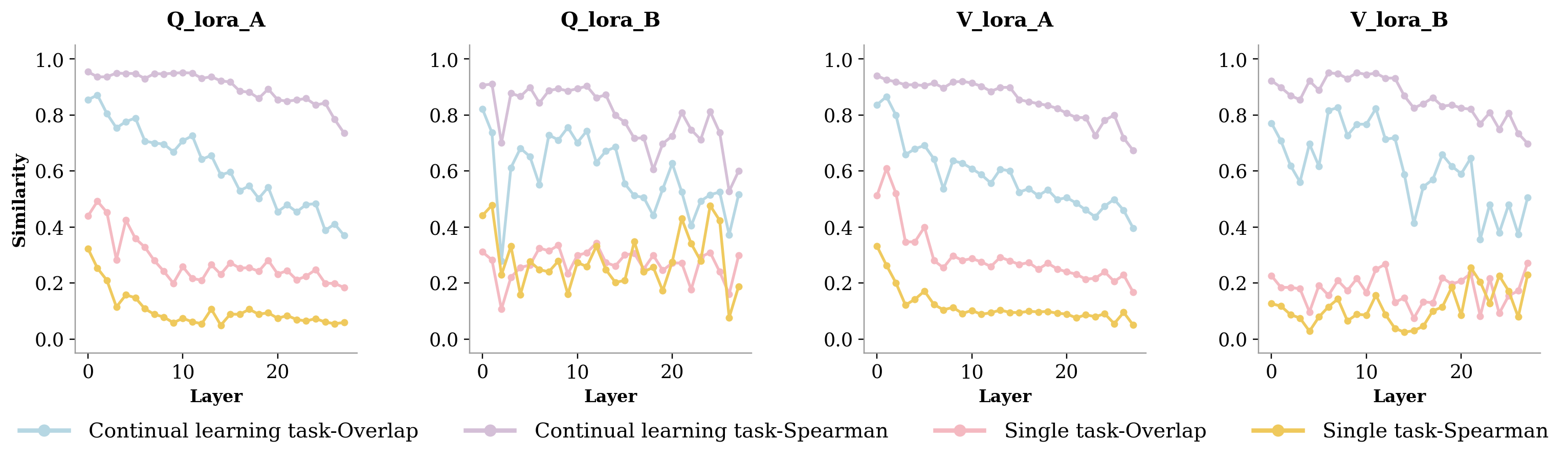}
  \caption{\small Similarity of task important parameters in single task and continual learning.}
  \label{fig:similarity}
\end{figure}

 Figure~\ref{fig:similarity} shows the results on Llama-3.2-Instruct-3B~\cite{grattafiori2024llama} with LoRA~\cite{hu2022lora} fine-tuning, where $\mathcal{T}_1$ is summary generation and $\mathcal{T}_2$ is code completion. In the independent single-task setting (pink and yellow), both the top-$K$ overlap and Spearman correlation remain low scores across layers, indicating that the important parameters vary across tasks. By contrast, after sequential continual learning (blue and purple), both similarity metrics increase. Consequently, updates for learning $\mathcal{T}_2$ are likely to modify parameters that are also critical to $\mathcal{T}_1$, thereby inducing interference and potential forgetting.

 \begin{figure}[htbp]
  \centering
  \includegraphics[width=\linewidth]{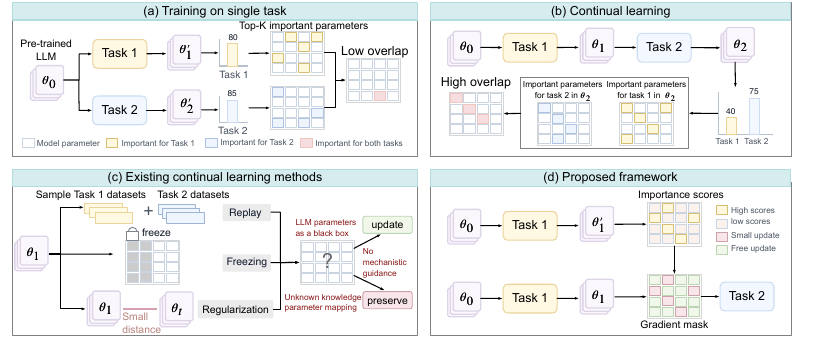}
  \caption{\small Motivation and overview of our proposed framework. (\textbf{a}): In single-task fine-tuning, a pre-trained LLM is separately adapted to Task 1 and Task 2, yielding task-specific models with low overlap among their top-$K$ important parameters. This suggests different tasks rely on distinct parameter subsets. (\textbf{b}): In continual learning, the important parameters for the two tasks become highly overlapped in the final model, indicating that learning Task 2 may modify parameters important to Task 1. (\textbf{c}): Existing methods lack a mechanistic understanding of how knowledge are distributed across LLM parameters. (\textbf{d}): Our framework obtain parameter importance and protects important parameters while allowing less important parameters to adapt to new tasks.} 
  \label{fig:method}
\end{figure}

Motivated by this observation, we propose an attribution-guided continual fine-tuning framework that mitigates forgetting by incorporating parameter importance. 
During continual learning, our method uses mechanistic attribution to find parameters important for old tasks. It keeps these important parameters stable, while allowing less important ones to change for learning new tasks. Experiments on LLMs under both full supervised fine-tuning and LoRA-based adaptation demonstrate that our framework consistently outperforms existing continual fine-tuning baselines. This shows that mechanistic attribution can serve as a principled signal for guiding continual fine-tuning in LLMs.

\section{Related work}
\textbf{Continual learning in LLMs}. Recent work on continual learning for LLMs adopts classical strategies to mitigate catastrophic forgetting, including regularization, replay, and parameter freezing. Among regularization-based methods, CLoRA~\cite{lu-etal-2025-controlled} regularizes LoRA updates in a subspace to reduce interference without replay. Replay-based methods mitigate forgetting by reusing or synthesizing old-task information. SSR~\cite{huang-etal-2024-mitigating} generates synthetic data with LLMs, then uses another LLM to refine and select high-quality samples for rehearsal. KPIG~\cite{he-etal-2025-dont} selects replay samples based on information gain. Freezing-based methods preserve prior knowledge by updating only a small subset of parameters or added modules. SAPT~\cite{zhao2024sapt} learns and selects parameter-efficient blocks via shared attentive learning and selection module. LoRAMoE~\cite{dou2024loramoe} freezes the backbone and adds MoE-style LoRA adapters, while this method~\cite{zheng2025spurious} freeze lower LLM layers and fine-tune upper layers. However, existing methods rarely analyze continual learning from the mechanistic perspective of LLMs, leaving the task-specific distribution of parameter importance largely unexplored.

\textbf{Explainability on LLMs}. Prior work has studied how input tokens contribute to LLM predictions. Layer-wise Relevance Propagation (LRP)~\cite{LRP} attributes predictions by backpropagating relevance scores to inputs, and has been adapted to Transformers~\cite{chefer2021transformer}. AttnLRP~\cite{attnlrp} further extends LRP to LLMs with attention-aware relevance propagation. Beyond input-token attribution, mechanistic interpretability studies how internal parameters and components shape model behavior. Key-value memory analyses of Transformer feed-forward networks~\cite{geva2021transformer} suggest how factual and linguistic information may be stored in parameters. Transformer circuit analysis~\cite{elhage2021mathematical} provides a framework for understanding component interactions. Edge-pruning methods~\cite{conmy2023towards} identify the most important connections in a Transformer computation graph to uncover circuits that drive model behavior. Howerve, existing methods lack a unified layer-wise relevance framework for tracking how input-token relevance is propagated across attention and feed-forward modules in each layer of LLMs.
\section{Background}
\textbf{Background on LLMs}. Large Language Models (LLMs) can be formulated as deep compositions of functions that progressively transform token representations~\cite{aubry2024transformer}. Given an input prompt tokenized as $x=[x_1,\dots,x_m]$, each token $x_i$ is mapped to a $d$-dimensional embedding and augmented with positional information, yielding the initial hidden state $h_i^{(0)}\in\mathbb{R}^{1\times d}$. The resulting sequence representation is
$
H^{(0)}=[h_1^{(0)};\dots;h_m^{(0)}]\in\mathbb{R}^{m\times d}$. 

The representations are then processed by $L$ Transformer blocks, each consisting of multi-head self-attention (MHA), a feed-forward network (FFN), layer normalization (LN), and residual connections. Let $H^{(l)}$ denote the hidden states after the $l$-th block. The $(l+1)$-th block is defined as
\begin{align}
f^{(l)} &= \mathrm{MHA}\big(\mathrm{LN}(H^{(l)})\big), 
\quad
g^{(l)} = \mathrm{LN}\big(H^{(l)} + f^{(l)}\big), 
\quad
H^{(l+1)} = H^{(l)} + f^{(l)} + \mathrm{FFN}\big(g^{(l)}\big).
\end{align}

After $L$ layers, $H^{(L)}$ are projected onto the vocabulary space to obtain the logits:
\begin{equation}
Z=\mathrm{LN}(H^{(L)})W_{\mathrm{vocab}}\in\mathbb{R}^{m\times |\mathcal{V}|},
\end{equation}
where $\mathcal{V}$ denotes the vocabulary, $|\mathcal{V}|$ is its size, and $W_{\mathrm{vocab}}\in\mathbb{R}^{d\times |\mathcal{V}|}$ is the output embedding matrix.
The next predicted token index is
$\hat{j}
=
\arg\max_{j\in\{1,\dots,|\mathcal{V}|\}} Z_{m,j}$,
which corresponds to the predicted token $\hat{x}_{m+1}=\mathcal{V}[\hat{j}]$.

\textbf{Background on continual learning}. In continual learning, the LLM is trained on a sequence of tasks $\{\mathcal{T}_1, \mathcal{T}_2, \ldots, \mathcal{T}_T\}$, where $T$ denotes the total number of tasks. We use $t \in \{1,\ldots,T\}$ to index the training stage and its corresponding task. Each task $\mathcal{T}_t$ is associated with a dataset  $
\mathcal{D}_t=\{(x^{(k)},y^{(k)})\}_{k=1}^{N_t}$,
where $x^{(k)}$ and $y^{(k)}$ denote the input prompt 
and target response of the $k$-th example, respectively. Let $\theta_t$ denote the model parameters after training on task $\mathcal{T}_t$. At stage $t$, the model is initialized from $\theta_{t-1}$ and optimized on the current dataset $\mathcal{D}_t$ using the autoregressive next-token prediction objective:
\begin{equation}
\mathcal{L}_t(\theta)
=
-\mathbb{E}_{(x^{(k)},y^{(k)})\sim \mathcal{D}_t}
\left[
\frac{1}{n}\sum_{j=1}^{n}
\log p_{\theta}\left(y^{(k)}_j \mid x^{(k)}, y^{(k)}_{<j}\right)
\right].
\end{equation}

\section{Method}
\label{sec:importance_scores}
We propose an importance-guided fine-tuning framework for continual learning in LLMs. For each task, our method first estimates element-wise importance scores for model parameters to measure their task-specific relevance (Section~\ref{sec:importance_scores}). 
During fine-tuning stage, parameters with high importance receive smaller updates, whereas those with low importance are updated freely. This allows the model to adapt to new tasks while reducing interference with previously acquired knowledge (Section~\ref{sec:fine_tuning}).

\subsection{Calculating element-wise importance of LLM parameters}
We apply the Layer-wise Relevance Propagation (LRP)~\cite{LRP} method to estimate the importance of LLM parameters in each layer. We first review the standard LRP formulation and its extension to Transformer architectures. Then, unlike prior applications that primarily attribute relevance to input features, we adapt LRP to quantify the contribution of parameters to the predicted next-token logit.

\subsubsection{LRP-based input attribution}
\textbf{Original LRP}. LRP attributes the prediction score to input neurons by propagating relevance backward through the network. For a neuron $a^{l+1} = \psi([a^{(l)}_1,\dots,a^{(l)}_n])$, where $\psi$ may be a
linear function or a composition of a linear function with
a nonlinear activation, its relevance $R^{l+1}$ is attributed to neurons in the preceding layer as: 
\begin{equation}
R^{(l)}_i = \frac{a^{(l
)}_i w^{l}_{i}}{\sum_{i^\prime} a^{(l)}_{i^\prime }w^{l}_{i^\prime}} R^{(l+1)}, 
\label{eq:lrp_local}
\end{equation}
where $w^{l}_{i}$ denotes weight from the neuron $a^{(l)}_i$ to the neuron $a^{(l+1)}$. 
For a multi-layer network, Eq.~(\ref{eq:lrp_local}) is recursively applied from the output logit $R^{(L)}$ to the input layer, producing input relevance scores $R^{(0)}_i$. The $R^{(0)}_i$ is computed as follows: 
\begin{equation}
R^{(0)}_i = \sum_{j,\dots,k}
\frac{a^{(0)}_i w^{0}_{i}}{\sum_{i^\prime} a^{(0)}_{i^\prime} w^{0}_{i^\prime}}
\cdots
\frac{a^{(L-1)}_k w^{L-1}_{k}}{\sum_{k^\prime} a^{(L-1)}_{k^\prime} w^{L-1}_{k^\prime}} 
\cdot
R^{(L)},\quad \sum_{i}R^{(0)}_i=R^{(L)}
\label{eq:lrp_multi}
\end{equation}

\textbf{LRP in Transformer}. The original LRP method can not used to multiplicative operations, such as 
$a^{(l+1)} = \sum_{i=1}^n a^{(l)}_i \cdot b^{(l)}_i$, when both $a^{(l)}_i$ and $b^{(l)}_i$ are inputs whose relevance needs to be attributed. 

However, in Transformer, matrix multiplications are extensively used in the self-attention module. To address this issue, 
AttnLRP~\cite{attnlrp} treats matrix multiplication as a bilinear operation: the relevance of $a^{(l)}_i$ is computed with $b^{(l)}_i$ fixed, and vice versa. Formally, given the output relevance score $R(a^{(l+1)})$,  the relevance assigned to $a^{(l)}_i$ and $b^{(l)}_i$, together with the conservation property, is given by:
\begin{equation}
\begin{aligned}
R(a^{(l)}_i)
=R(b^{(l)}_i)=
\frac{1}{2}
\frac{a^{(l)}_i b^{(l)}_i}
{\sum_{j=1}^{n} a^{(l)}_j b^{(l)}_j }
R(a^{(l+1)}), \quad \sum_{i=1}^{n} R(a^{(l)}_i) + \sum_{i=1}^{n} R(b^{(l)}_i)
=
R(a^{(l+1)})
\end{aligned}
\label{eq:attnlrp_bilinear}
\end{equation}

\subsubsection{LRP-based parameters attribution in LLMs}
\label{sec:lrp_attribution_llms}
Existing LRP methods, including their Transformer extensions, primarily attribute predictions to inputs, while leaving model parameters unattributed. We adopt the bilinear attribution principle of AttnLRP to the FFN and MHA modules and derive parameter-level relevance scores for each LLM layer to the next token logit $Z_{m,\hat{j}}$, where $\hat{j}
=\arg\max_{j\in\{1,\dots,|\mathcal{V}|\}} Z_{m,j}$. 
These relevance scores are then used to guide fine-tuning process.
\begin{lemma}
\label{thm:lrp_ffn}
For a linear transformation $C = PW+B$, where 
$P\in\mathbb{R}^{m\times d_{\mathrm{in}}}$, 
$W\in\mathbb{R}^{d_{\mathrm{in}}\times d_{\mathrm{out}}}$, and 
$B, C\in\mathbb{R}^{m\times d_{\mathrm{out}}}$. Let $\odot$ denote element-wise multiplication, and the division is applied element-wise. Given the output relevance $R(C)\in\mathbb{R}^{m\times d_{\mathrm{out}}}$, the relevance assigned to $P$, $W$, and $B$ under the bilinear LRP rule is obtained from the following formula and satisfies the conservation property: 
\begin{equation*}
\begin{aligned}
R(P)
=
\frac{1}{2}
P\odot
\left[
\left(\frac{R(C)}{C}\right)W^\top
\right],
R(W)
=
\frac{1}{2}
W\odot
\left[
P^\top
\left(\frac{R(C)}{C}\right)
\right],
R(B)
=
B\odot
\frac{R(C)}{C}, 
\end{aligned}
\label{eq:linear_param_relevance_matrix}
\end{equation*}
\begin{equation}
\sum R(P)+\sum R(W)+\sum R(B)=\sum R(C),
\label{eq:linear_relevance_conservation}
\end{equation}
\end{lemma}
Lemma~\ref{thm:lrp_ffn} quantifies the contributions of input activations, model parameters, and biases in matrix multiplication. We apply this result to derive parameter-level relevance scores for the FFN and MHA modules in LLMs.

\textbf{LRP Attribution in FFN}. Let $\Delta_{\mathrm{FFN}}^{(l)}=\mathrm{FFN}\big(g^{(l)}\big)$ and the FFN module is defined as
\[
U^{(l)} = g^{(l)} W_1 + b_1,
\quad
S^{(l)} = \sigma(U^{(l)}),
\quad
\Delta_{\mathrm{FFN}}^{(l)} = S^{(l)} W_2 + b_2,
\]
where $W_1\in\mathbb{R}^{d\times d_{\mathrm{ff}}}$, $b_1\in\mathbb{R}^{d_{\mathrm{ff}}}$, $W_2\in\mathbb{R}^{d_{\mathrm{ff}}\times d}$, and $b_2\in\mathbb{R}^{d}$. The intermediate activations satisfy $U^{(l)},S^{(l)}\in\mathbb{R}^{m\times d_{\mathrm{ff}}}$, and the FFN output satisfies $\Delta_{\mathrm{FFN}}^{(l)}\in\mathbb{R}^{m\times d}$. We then apply Lemma~\ref{thm:lrp_ffn} sequentially to the FFN, yielding the following relevance propagation rules for the input and parameters.

\begin{proposition}[FFN relevance propagation]
\label{prop:ffn_relevance}
Given the output relevance $R(\Delta_{\mathrm{FFN}}^{(l)})$, and treating the activation $\sigma(\cdot)$ as identity
relevance propagation , i.e., $R(U^{(l)})=R(S^{(l)})$. $\odot$ denotes element-wise multiplication, division is applied element-wise, and $\mathbf{1}_m\in\mathbb{R}^{m\times 1}$ is an all-one vector. The relevance scores of the FFN input and parameters are given by
\begin{align}
R(W_2^{(l)})
&=
\frac{1}{2}
W_2^{(l)}\odot
\left[
\left(S^{(l)}\right)^\top
\left(
\frac{R(\Delta_{\mathrm{FFN}}^{(l)})}
{\Delta_{\mathrm{FFN}}^{(l)}}
\right)
\right],
\quad
R(b_2^{(l)})
=
b_2^{(l)}
\odot
\mathbf{1}_m^\top
\left(
\frac{R(\Delta_{\mathrm{FFN}}^{(l)})}
{\Delta_{\mathrm{FFN}}^{(l)}}
\right), \notag \\
R(S^{(l)})
&=
\frac{1}{2}
S^{(l)}\odot
\left[
\left(
\frac{R(\Delta_{\mathrm{FFN}}^{(l)})}
{\Delta_{\mathrm{FFN}}^{(l)}}
\right)
\left(W_2^{(l)}\right)^\top
\right],
\quad
R(g^{(l)})
=
\frac{1}{2}
g^{(l)}\odot
\left[
\left(
\frac{R(S^{(l)})}
{U^{(l)}}
\right)
\left(W_1^{(l)}\right)^\top
\right], 
\notag \\
R(W_1^{(l)})
&=
\frac{1}{2}
W_1^{(l)}\odot
\left[
\left(g^{(l)}\right)^\top
\left(
\frac{R(S^{(l)})}
{U^{(l)}}
\right)
\right],
\quad
R(b_1^{(l)})
=
b_1^{(l)}
\odot
\mathbf{1}_m^\top
\left(
\frac{R(S^{(l)})}
{U^{(l)}}
\right). 
\end{align}
\end{proposition}
Proposition~\ref{prop:ffn_relevance} provides relevance scores for both inputs and parameters in LLM FFN modules. These scores further satisfy the following conservation property:
\begin{equation*}
\sum R(g^{(l)})
+\sum R(W_1^{(l)})
+\sum R(b_1^{(l)})
+\sum R(W_2^{(l)})
+\sum R(b_2^{(l)})
=
\sum R(\Delta_{\mathrm{FFN}}^{(l)}).
\label{eq:ffn_relevance_conservation}
\end{equation*}

\textbf{LRP Attribution in MHA}. 
Let $X^{(l)}=\mathrm{LN}(H^{(l)})$, where $H^{(l)},X^{(l)}\in\mathbb{R}^{m\times d}$. 
Let $d_{\mathrm{head}}$ denote the dimension of each attention head. 
For $r=1,\dots,d/d_{\mathrm{head}}$, the $r$-th head is computed as
\begin{equation*}
\begin{aligned}
Q_r^{(l)} &= X^{(l)}W_{Q,r}^{(l)}, \quad
K_r^{(l)} = X^{(l)}W_{K,r}^{(l)}, \quad
V_r^{(l)} = X^{(l)}W_{V,r}^{(l)}, \quad
A_r^{(l)} =
\mathrm{softmax}\!\left(
\frac{Q_r^{(l)}(K_r^{(l)})^\top}{\sqrt{d_{\mathrm{head}}}}
\right), \\
O_r^{(l)} &= A_r^{(l)}V_r^{(l)}, \quad
O^{(l)}=\mathrm{Concat}\left(O_1^{(l)},\dots,O_{d/d_{\mathrm{head}}}^{(l)}\right),
\quad
f^{(l)}=O^{(l)}W_O^{(l)}.
\end{aligned}
\end{equation*} 
Where $W_{Q,r}^{(l)},W_{K,r}^{(l)},W_{V,r}^{(l)}\in\mathbb{R}^{d\times d_{\mathrm{head}}}$, 
$Q_r^{(l)},K_r^{(l)},V_r^{(l)},O_r^{(l)}\in\mathbb{R}^{m\times d_{\mathrm{head}}}$,  
$A_r^{(l)}\in\mathbb{R}^{m\times m}$, $W_O^{(l)}\in\mathbb{R}^{d\times d}$ and $f^{(l)}\in\mathbb{R}^{m\times d}$. Applying Lemma~\ref{thm:lrp_ffn} to the MHA module gives the following relevance propagation rules for its inputs and parameters.
\begin{proposition}[MHA relevance propagation]
\label{prop:mha_relevance}
Given the output relevance $R(f^{(l)})$ of the MHA module, let $[\cdot]_r$ denote the $r$-th column-wise block corresponding to the $r$-th attention head. 
We treat the softmax operation as identity relevance propagation, $\Lambda_f^{(l)}=R(f^{(l)})/f^{(l)}$,
$R(O_r^{(l)})=\left[\frac{1}{2}O^{(l)}\odot\left(\Lambda_f^{(l)}(W_O^{(l)})^\top\right)\right]_r$, 
$\Lambda_{O,r}^{(l)}=R(O_r^{(l)})/O_r^{(l)}$, and 
$\Phi_r^{(l)}=
\left(A_r^{(l)}\odot\left[\Lambda_{O,r}^{(l)}(V_r^{(l)})^\top\right]\right)/E_r^{(l)}$. 
We further define 
$\Theta_{Q,r}^{(l)}=\Phi_r^{(l)}K_r^{(l)}$, 
$\Theta_{K,r}^{(l)}=(\Phi_r^{(l)})^\top Q_r^{(l)}$, and 
$\Theta_{V,r}^{(l)}=(A_r^{(l)})^\top\Lambda_{O,r}^{(l)}$. Let $\alpha=\frac{1}{8\sqrt{d_{\mathrm{head}}}}$
All divisions are applied element-wise. 
Then the parameters and onput relevance scores of the MHA module are:
\begin{align}
R(W_O^{(l)})
&=
\frac{1}{2}
W_O^{(l)}
\odot
\left[
\left(O^{(l)}\right)^\top
\Lambda_f^{(l)}
\right],
\quad
R(W_{Q,r}^{(l)})
=
\alpha
W_{Q,r}^{(l)}
\odot
\left[
\left(X^{(l)}\right)^\top
\Theta_{Q,r}^{(l)}
\right], \notag \\
R(W_{V,r}^{(l)})
&=
\frac{1}{4}
W_{V,r}^{(l)}
\odot
\left[
\left(X^{(l)}\right)^\top
\Theta_{V,r}^{(l)}
\right],
\quad
R(W_{K,r}^{(l)})
=
\alpha
W_{K,r}^{(l)}
\odot
\left[
\left(X^{(l)}\right)^\top
\Theta_{K,r}^{(l)}
\right], \notag \\
R(X^{(l)})
&=
X^{(l)}
\odot
\sum_{r=1}^{d/d_{\mathrm{head}}}
\left[
\alpha
\Theta_{Q,r}^{(l)}
\left(W_{Q,r}^{(l)}\right)^\top
+
\alpha
\Theta_{K,r}^{(l)}
\left(W_{K,r}^{(l)}\right)^\top
+
\frac{1}{4}
\Theta_{V,r}^{(l)}
\left(W_{V,r}^{(l)}\right)^\top
\right].
\label{eq:mha_relevance_compact}
\end{align}
\end{proposition}
Moreover, these relevance scores satisfy:
\begin{equation*}
\sum R(X^{(l)})
+
\sum R(W_O^{(l)})
+
\sum_{r=1}^{d/d_{\mathrm{head}}}
\left[
\sum R(W_{Q,r}^{(l)})
+
\sum R(W_{K,r}^{(l)})
+
\sum R(W_{V,r}^{(l)})
\right]
=
\sum R(f^{(l)}). 
\label{eq:mha_relevance_conservation}
\end{equation*}
\textbf{LoRA-based attribution}.   
For LoRA-adapted MHA, we write each projection as 
$\widetilde{W}_{O}^{(l)}=W_{O}^{(l)}+A_{O}^{(l)}B_{O}^{(l)}$ and 
$\widetilde{W}_{\star,r}^{(l)}=W_{\star,r}^{(l)}+A_{\star,r}^{(l)}B_{\star,r}^{(l)}$, 
where $\star\in\{Q,K,V\}$, and $A,B$ are trainable LoRA parameters. 
We compute relevance scores for the frozen backbone parameters, the trainable LoRA parameters and the input.

\textbf{LRP Attribution in Transformer Block.} 
We initialize the output relevance using the predicted next-token logit,
$R(H^{(L)})=Z_{m,\hat{j}}$, where $\hat{j}
=\arg\max_{j\in\{1,\dots,|\mathcal{V}|\}} Z_{m,j}$.
We then extend relevance propagation from individual FFN and MHA modules to the standard Transformer block, as illustrated in Figure~\ref{fig:method}. 
Figure~\ref{fig:method}(a) shows the standard forward pass in Transformer block in LLMs. 
Figure~\ref{fig:method}(b) presents our relevance propagation procedure, which decomposes the output relevance $R(H^{(l+1)})$ across the residual, FFN, and MHA branches, assigning relevance scores to both intermediate activations and model parameters. 
Given $R(H^{(l+1)})$, the residual connection first distributes relevance to its additive inputs:
\begin{equation*}
R(H^{(l+1)})
=
R(H^{(l)})
+
R(f^{(l)})
+
R\big(\mathrm{FFN}(g^{(l)})\big).
\label{eq:block_residual_split}
\end{equation*}

The $R\big(\mathrm{FFN}(g^{(l)})\big)$ is then propagated through the FFN module using the proposition~\ref{prop:ffn_relevance}:
\begin{equation*}
R(\mathrm{FFN}(g^{(l)}))
\;\rightarrow\;
\big\{
R(g^{(l)}),\;
R(W_1^{(l)}),\;
R(W_2^{(l)}),\;
R(b_1^{(l)}),\;
R(b_2^{(l)})
\big\}. 
\label{eq:block_ffn_full}
\end{equation*}
We treat layer normalization as identity relevance propagation, yielding 
$R(g^{(l)})
=
R(H^{(l)})
+
R(f^{(l)})$. Then $R(f^{(l)})$ is propagated through the MHA module, yielding both input and parameter relevance using the proposition~\ref{prop:mha_relevance}:
\begin{equation*}
R(f^{(l)})
\;\rightarrow\;
\big\{
R(X^{(l)}),\;
R(W_O^{(l)}),\;
\{R(W_{Q,r}^{(l)}), R(W_{K,r}^{(l)}), R(W_{V,r}^{(l)})\}_{r}
\big\},
\label{eq:block_mha_full}
\end{equation*}
where $X^{(l)}=\mathrm{LN}(H^{(l)})$, and $R(X^{(l)})=R(H^{(l)})$. 

Combining all branches, we obtain relevance scores for both the input $H^{(l)}$ and all model parameters in the $l$-th Transformer block. 
We denote the parameter set by $\mathcal{W}^{(l)}=\Big\{
W_O^{(l)},
\{W_{Q,r}^{(l)}, W_{K,r}^{(l)}, W_{V,r}^{(l)}\}_{r},
W_1^{(l)}, W_2^{(l)},
b_1^{(l)}, b_2^{(l)}\Big\}$, and their corresponding relevance scores by $R(\mathcal{W}^{(l)})
=
\Big\{
R(W_O^{(l)}),
\{R(W_{Q,r}^{(l)}), R(W_{K,r}^{(l)}), R(W_{V,r}^{(l)})\}_{r},
R(W_1^{(l)}), R(W_2^{(l)}),
R(b_1^{(l)}), R(b_2^{(l)})
\Big\}$. $R(\mathcal{W}^{(l)})$ is further used to guide parameter updates in the continual learning fine-tuning process.

\begin{figure}[htbp]
  \centering
  \includegraphics[width=\linewidth]{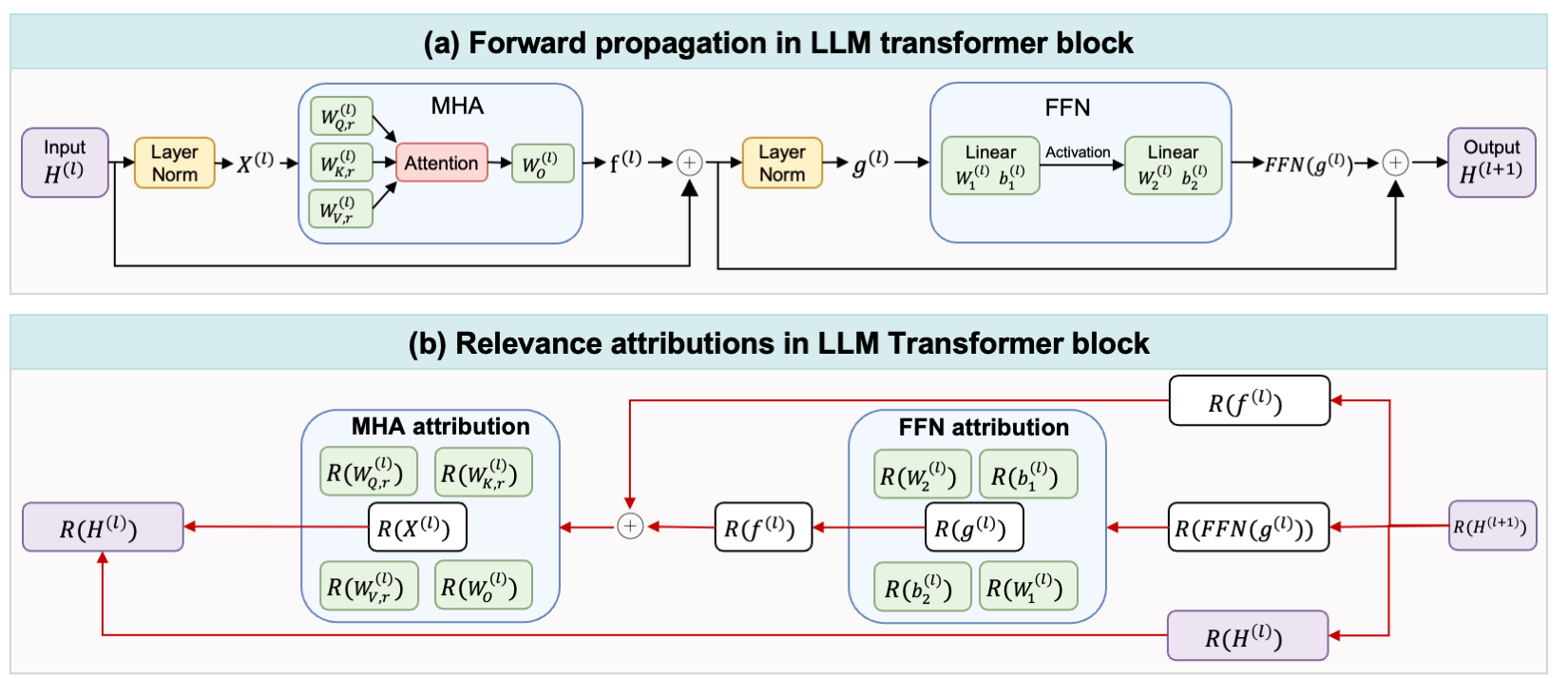}
  \caption{\small Forward propagation and LRP-based parameter Attribution in an LLM Transformer Block.}
  \label{fig:method}
\end{figure}

\subsection{Attribution guided fine-tuning for continual learning}
\label{sec:fine_tuning}
\textbf{Importance prior estimation for each task}. 
For each task, we first estimate parameter importance in a single-task setting and use it as a prior for continual learning. Given task $\mathcal{T}_t$ with dataset $\mathcal{D}_t=\{(x^{(k)},y^{(k)})\}_{k=1}^{N_t}$, we train a task-specific model with parameters $\theta_t'$. Let $\mathcal{W}^{(l,t)}$ denote the parameters of the $l$-th Transformer block, including attention projections, feed-forward weights, and biases. To obtain reliable attribution signals, we only consider samples correctly generated by the task-specific model $\mathcal{C}_t$ and normalize each sample-level relevance map as
\begin{equation}
\widehat{R}^{(k)}(W^{(l,t)})
=
\operatorname{sign}\!\left(R^{(k)}(W^{(l,t)})\right)
\odot
\frac{
\log\!\left(1+\left|R^{(k)}(W^{(l,t)})\right|\right)
}{
\max \log\!\left(1+\left|R^{(k)}(W^{(l,t)})\right|\right)+\epsilon
}.
\label{eq:correct_samples}
\end{equation}
Let $\mathcal{K}_{u}(W^{(l,t)})
=
\operatorname{TopK}_{k\in\mathcal{C}_t}
\left(
\widehat{R}^{(k)}(W^{(l,t)})[u]
\right)$ and we aggregate sample-level attributions into a task-level importance prior by retaining the top-$K$ normalized relevance values :
\begin{equation}
R(W^{(l,t)})[u]
=
\frac{1}{K}
\sum_{k\in\mathcal{K}_{u}(W^{(l,t)})}
\widehat{R}^{(k)}(W^{(l,t)})[u].
\label{eq:topk_indices}
\end{equation}
Where $\operatorname{TopK}$ returns the indices of the $K$ largest values. The element-wise importance  $R(W^{(l,t)})$ quantifies which parameters are important in  task $\mathcal{T}_t$.

\textbf{Attribution-guided continual learning}
During continual learning, for the task $t$, the $R(W^{(l,1)}),\dots,R(W^{(l,t-1)})$ are used to protect parameters that are critical to previously learned tasks. When learning the current task $\mathcal{T}_t$, we summarize historical importance for each parameter by taking the element-wise maximum over all prior tasks:
\begin{equation}
\bar{R}(W^{(l,t)})
=
\left[
\max_{\tau<t} R(W^{(l,\tau)})
\right]_+,
\quad
\bar{R}(W^{(l,t)})\in[0,1],
\label{eq:historical_importance}
\end{equation}
where $[x]_+=\max(x,0)$ is applied element-wise. We incorporate this prior by modulating the gradient during fine-tuning:
\begin{equation}
\widetilde{\nabla}_{W^{(l,t)}}\mathcal{L}_t
=
\left(
\mathbf{1}
-
\bar{R}(W^{(l,t)})
\right)
\odot
\nabla_{W^{(l,t)}}\mathcal{L}_t.
\label{eq:importance_guided_gradient}
\end{equation} 
Here, $(\mathbf{1}-\bar{R}(W^{(l,t)}))$ serves as an element-wise gradient gate. 
Parameters that are important to any previous task have large historical relevance, making the gate close to zero and thus receiving smaller updates. 
Parameters with low historical relevance keep gates close to one, so their gradients are largely unchanged when learning the current task.

\section{Experiments}
\textbf{Datasets}.  We evaluate our method on two benchmark datasets. The first is Concept-1K~\cite{zheng2024concept}, which consists of 10 diverse tasks designed to assess the model’s ability to learn and generalize across different concepts. The second is TRACE, from which we select seven representative tasks: FOMC, MeetingBank, PY150, ScienceQA, NumGLUE-cm, NumGLUE-ds, and 20Minute. These tasks cover a broad range of scenarios, including financial text understanding, meeting summarization, code completion, scientific question answering, numerical reasoning, and long-context comprehension. 

\textbf{Baselines}. 
We compare attribution-guided fine-tuning (AGFT) with four continual learning baselines: \textbf{SEQ}, which sequentially fine-tunes the model on each task; \textbf{EWC}~\cite{ewc}, a regularization-based method that estimates parameter importance using the diagonal Fisher information matrix; \textbf{Replay}, which trains on new-task data together with a buffer of previous examples; and \textbf{Freeze}~\cite{zheng2025spurious}, which freezes the bottom Transformer layers during training.

\textbf{Continual learning setting and models}.
We first independently train a model for each task and estimate parameter importance using our attribution method. During sequential continual learning, the estimated importance scores from previous tasks are used to guide model updates. Experiments are conducted on GPT-2~\cite{radford2019language} and Llama-3.2-3B-Instruct, where GPT-2 is trained via full-parameter fine-tuning on Concept-1K, and Llama-3.2-3B-Instruct is trained with LoRA on both Concept-1K and TRACE.
\begin{table}[t]
\centering
\caption{Performance comparison of different continual learning methods on the Concept-1K dataset using GPT-2 and Llama-3.2-3B-Instruct as backbone models. Best results are highlighted in bold.}
\label{tab:results_concept-1k}
\small
\setlength{\tabcolsep}{2.7pt}

\sisetup{
    detect-weight=true,
    detect-inline-weight=math,
    table-number-alignment=center,
    table-format=2.2
}
\begin{tabular}{
    ll
    *{11}{S}
}
\toprule
\textbf{Model} & \textbf{Method}
& {\textbf{Task1}} & {\textbf{Task2}} & {\textbf{Task3}} & {\textbf{Task4}}
& {\textbf{Task5}} & {\textbf{Task6}} & {\textbf{Task7}} & {\textbf{Task8}}
& {\textbf{Task9}} & {\textbf{Task10}} & {\textbf{Average}} \\
\midrule

\multirow{5}{*}{Llama}
& \textbf{EWC}    & 12.43 & 13.40 & 11.50 & 14.15 & 12.72 & 13.23 & 12.78 & 14.28 & 12.81 & 16.00 & 13.33 \\
& \textbf{SEQ}    & 11.78 & 14.88 & 12.48 & 13.66 & 13.84 & 14.09 & 12.66 & 13.74 & 14.85 & 15.57 & 13.75 \\
& \textbf{Replay} & 13.96 & 15.64 & 12.90 & 13.78 & 12.66 & 14.15 & 13.32 & 13.57 & 12.16 & 13.63 & 13.58 \\
& \textbf{Freeze} & 13.07 & 12.63 & 12.23 & 14.57 & 13.49 & 14.09 & 12.78 & 13.63 & 13.59 & 13.56 & 13.37 \\
& \cellcolor{gray!10}\textbf{AGFT}
& \cellcolor{gray!10}\bfseries 16.31
& \cellcolor{gray!10}\bfseries 17.30
& \cellcolor{gray!10}\bfseries 15.40
& \cellcolor{gray!10}\bfseries 16.60
& \cellcolor{gray!10}\bfseries 16.08
& \cellcolor{gray!10}\bfseries 16.32
& \cellcolor{gray!10}\bfseries 15.90
& \cellcolor{gray!10}\bfseries 16.41
& \cellcolor{gray!10}\bfseries 15.93
& \cellcolor{gray!10}\bfseries 17.28
& \cellcolor{gray!10}\bfseries 16.35 \\

\midrule

\multirow{5}{*}{GPT-2}
& \textbf{EWC}    & 0.00 & 0.00 & 0.06 & 0.12 & 0.06 & 0.12 & 0.06 & 0.24 & 0.06 & 0.24 & 0.10 \\
& \textbf{SEQ}    & 3.71 & 5.02 & 5.48 & 4.78 & 5.77 & 4.51 & 5.76 & 6.75 & 6.35 & 10.58 & 5.87 \\
& \textbf{Replay} & 8.48 & \bfseries 9.03 & 7.06 & 7.04 & 7.19 & 8.96 & \bfseries 7.14 & \bfseries 8.00 & \bfseries 5.69 & 7.12 & 7.57 \\
& \textbf{Freeze} & 4.54 & 5.08 & 5.30 & 5.08 & 5.83 & 5.75 & 5.64 & 6.81 & 6.35 & 11.38 & 6.17 \\
& \cellcolor{gray!10}\textbf{AGFT}
& \cellcolor{gray!10}\bfseries 9.78
& \cellcolor{gray!10} 7.38
& \cellcolor{gray!10}\bfseries 7.85
& \cellcolor{gray!10}\bfseries 7.72
& \cellcolor{gray!10}\bfseries 7.48
& \cellcolor{gray!10}\bfseries 10.20
& \cellcolor{gray!10}\bfseries 7.14
& \cellcolor{gray!10} 6.99
& \cellcolor{gray!10}\bfseries 5.69
& \cellcolor{gray!10}\bfseries 9.00
& \cellcolor{gray!10}\bfseries 7.92 \\

\bottomrule
\end{tabular}
\end{table}

\textbf{Evaluation metrics}. We evaluate model performance using different metrics for dfferent task type, including Accuracy, ROUGE-L, Edit Similarity, and SARI. Accuracy is computed by exact match for Llama-3.2-3B-Instruct and by answer containment for GPT-2. 
ROUGE-L, Edit Similarity, and SARI are used for summarization, code completion, and text simplification tasks, respectively.

\textbf{Performance}. 
Table~\ref{tab:results_concept-1k} report the continual learning results. Overall, AGFT achieves the best average performance across both datasets and backbone models. On Concept-1K, AGFT consistently improves all baselines with Llama-3.2-3B-Instruct in each task. With GPT-2, it also obtains the best results on 9 of 10 tasks.
 In TRACE dataset, it achieves the best result on 5 out of 7 tasks. These results demonstrate the effectiveness of AGFT in continual learning.
EWC performs the worst because Fisher information is poorly estimated in LLMs, leading to unreliable parameter-importance estimates. As a result, EWC fails to identify critical parameters and cannot effectively balance adaptation and forgetting.
Freeze-based methods cannot precisely control individual parameters. Freezing whole layers can block useful updates for new tasks while still failing to fully preserve prior knowledge. Replay method can reduce forgetting because the model has limited capacity and benefits from repeatedly seeing past-task data in GPT-2. For stronger models Llama-3.2-3B-Instruct, replay is less useful because model learn more transferable task knowledge rather than memorization.

\begin{table}[t]
\centering
\caption{Performance comparison of different continual learning methods on the TRACE dataset using Llama-3.2-Instruct as the backbone model. Best results are highlighted in bold.}
\label{tab:llama_trace_results}
\small
\setlength{\tabcolsep}{7pt}

\sisetup{
    detect-weight=true,
    detect-inline-weight=math,
    table-number-alignment=center,
    table-format=2.2
}

\begin{tabular}{
    ll
    *{8}{S}
}
\toprule
\textbf{Model} & \textbf{Method}
& {\textbf{Task1}}
& {\textbf{Task2}}
& {\textbf{Task3}}
& {\textbf{Task4}}
& {\textbf{Task5}}
& {\textbf{Task6}}
& {\textbf{Task7}}
& {\textbf{Average}} \\
\midrule

\multirow{5}{*}{Llama}
& \textbf{EWC}
& 48.39 & 0.00 & 1.00 & 16.00 & 46.43 & \bfseries 46.95 & 0.00 & 22.68 \\
& \textbf{SEQ}
& 43.15 & 33.50 & \bfseries 32.92 & 21.00 & 40.74 & 44.31 & 39.59 & 36.45 \\
& \textbf{Replay}
& 40.60 & 32.69 & 30.80 & 13.00 & 37.00 & 40.00 & 39.44 & 33.36 \\
& \textbf{Freeze}
& 42.54 & 32.49 & 23.59 & 20.00 & 29.63 & 44.92 & 38.22 & 33.05 \\
& \cellcolor{gray!10}\textbf{AGFT}
& \cellcolor{gray!10}\bfseries 57.66
& \cellcolor{gray!10}\bfseries 50.50
& \cellcolor{gray!10} 30.78
& \cellcolor{gray!10}\bfseries 22.00
& \cellcolor{gray!10}\bfseries 49.38
& \cellcolor{gray!10} 43.69
& \cellcolor{gray!10}\bfseries 39.93
& \cellcolor{gray!10}\bfseries 41.99 \\

\bottomrule
\end{tabular}
\end{table} 

\begin{figure}[htbp]
  \centering
  \includegraphics[width=\linewidth]{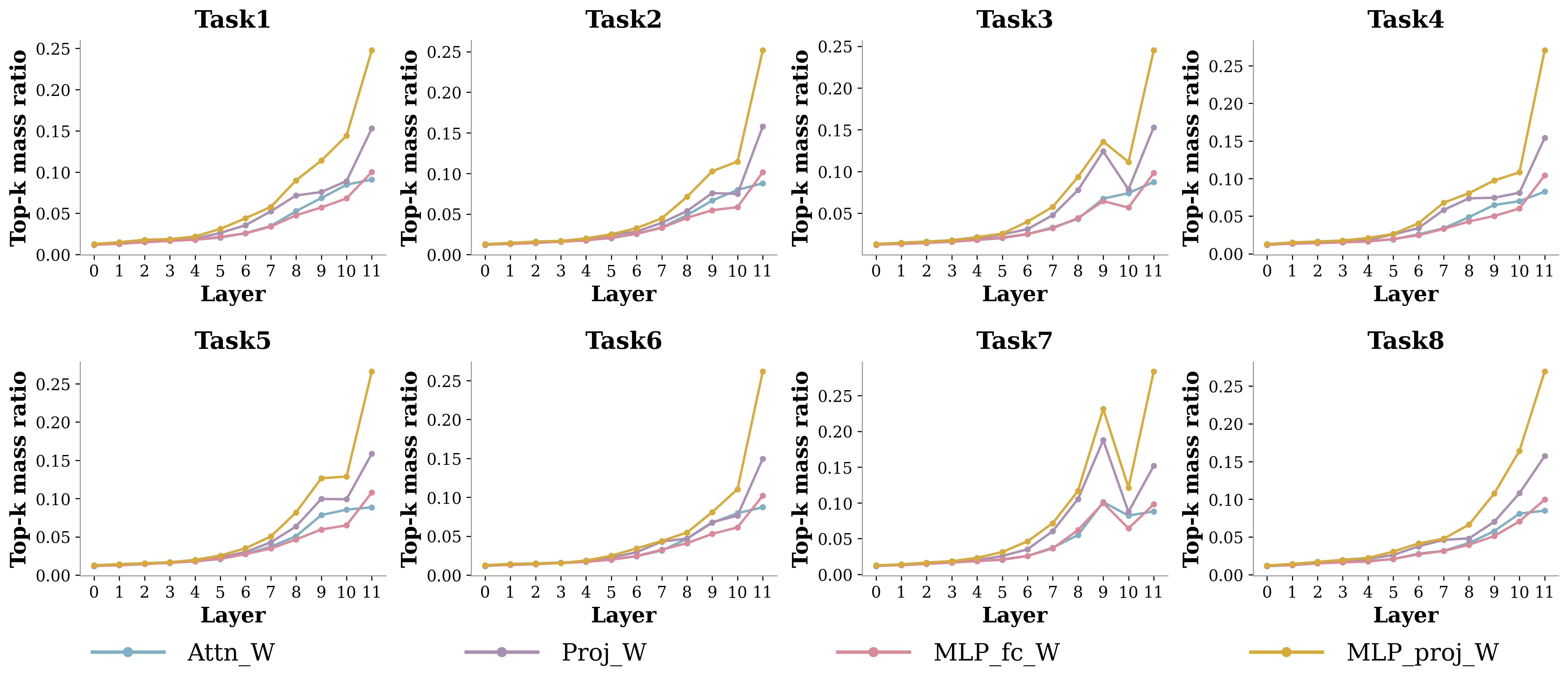}
  \caption{\small Top-$k$ mass ratio across layers for different matrices and tasks in GPT-2.}
  \label{fig:gpt2_explanation}
\end{figure}

\begin{figure}[htbp]
  \centering
  \includegraphics[width=\linewidth]{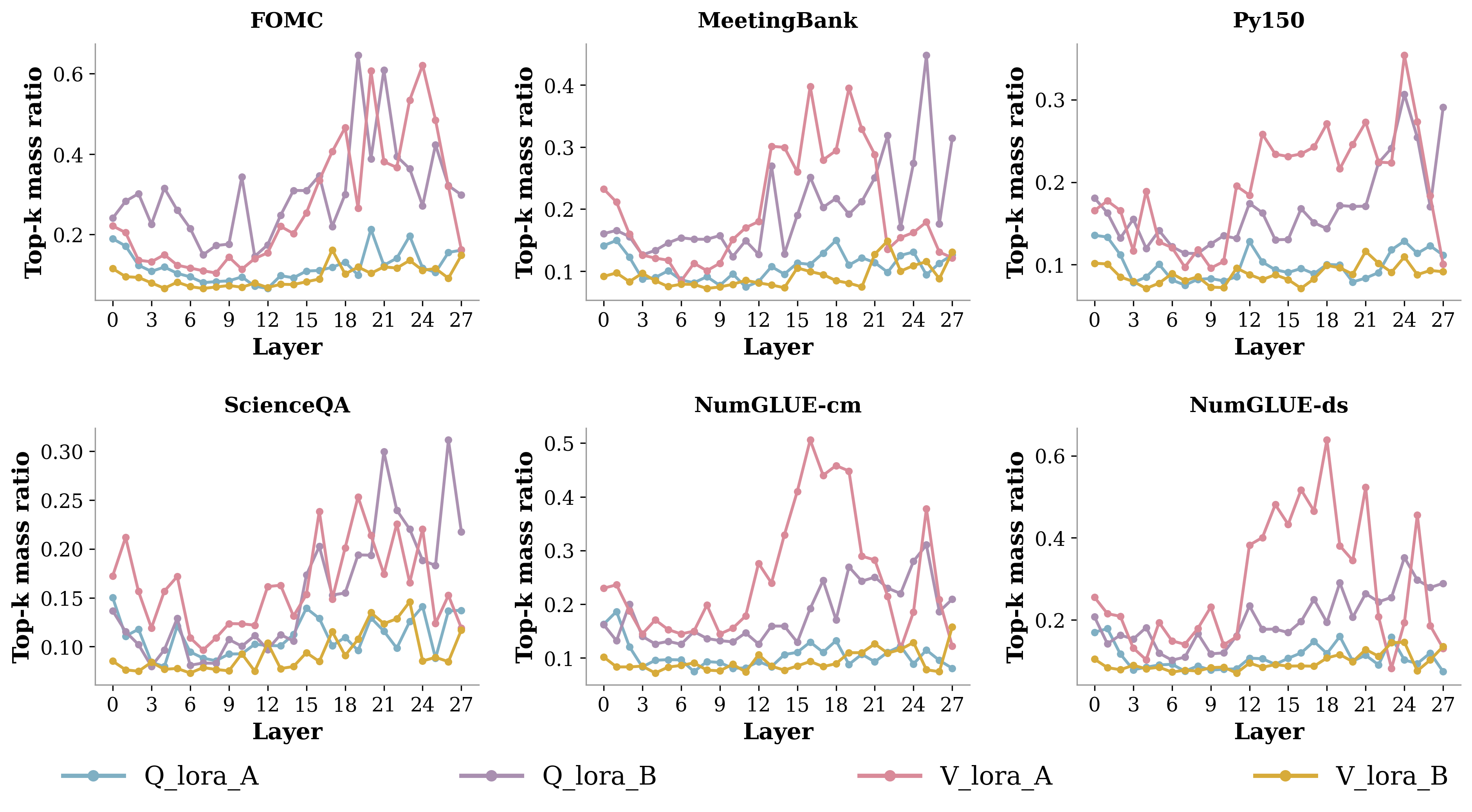}
  \caption{\small Top-$k$ mass ratio across layers for different LoRA matrices and tasks in Llama-3.2-Instruct-3B.}
  \label{fig:llama_explanation}
\end{figure}

\textbf{Case study on concentration of importance distributions in LLMs.}
For each layer $l$ and parameter matrix $W^{(l)}$, we compute the importance score $\widehat{R}^{(k)}(W^{(l,t)})$ following Eq.~(\ref{eq:correct_samples}). 
To measure how concentrated the importance distribution is, we define the Top-$k$ mass ratio for each matrix in $\mathcal{T}_t$ as
$
\frac{
\operatorname{TopK}_{k\in\mathcal{C}_t}
\left(
|\widehat{R}^{(k)}(W^{(l,t)})[u]|
\right)
}{
\sum_u |\widehat{R}^{(k)}(W^{(l,t)})[u]|
}$.
A larger value indicates that importance is concentrated on a small subset of parameters, whereas a smaller ratio suggests a more diffuse distribution.
We use this metric to compare concentration patterns across layers, matrices, and models.
\textbf{Observations on GPT-2.}
As shown in Figure~\ref{fig:gpt2_explanation}, the later transformer blocks rely on a smaller subset of parameters in GPT-2, with the degree of concentration varying across tasks. It suggests that task-specific knowledge maybe encoded in a sparse manner in the deeper part of the network. \textbf{Observations on Llama-3.2-Instruct-3B.}
In Figure~\ref{fig:llama_explanation}, for $V_{\mathrm{lora\_A}}$ and $Q_{\mathrm{lora\_B}}$, the ratio remains low in lower layers but increases in upper layers, while $Q_{\mathrm{lora\_A}}$ and $V_{\mathrm{lora\_B}}$ stay flat. 
This suggests that deeper layers tend to concentrate task-relevant updates on a small subset of parameters, but only for $V_{\mathrm{lora\_A}}$ and $Q_{\mathrm{lora\_B}}$. 
Overall, the results reveal a sparse pattern of parameter importance, supporting the idea that protecting a small number of critical parameters can help preserve previous tasks during continual learning.

\section{Conclusions and Limitations}
\label{sec:conclusion_limitation}
We study continual fine-tuning in LLMs, where existing methods overlook the internal mechanisms underlying LLM adaptation. Thus, we propose AGFT, which leverages LRP to estimate parameter importance across different modules during model propagation. These importance scores serve as model-intrinsic prior knowledge to guide continual fine-tuning. Extensive experiments demonstrate that AGFT consistently outperforms strong baseline methods. The limitation of our method is that estimating parameter importance introduces additional computational overhead. 

\newpage
\bibliography{example}
\bibliographystyle{unsrt}


\end{document}